%% file: main.tex
\documentclass[10pt,twocolumn,letterpaper]{article}

\usepackage{cvpr}              

\input{preamble}

\definecolor{cvprblue}{rgb}{0.21,0.49,0.74}
\usepackage[pagebackref,breaklinks,colorlinks,allcolors=cvprblue]{hyperref}
\def\paperID{SAIAD-8} 
\def\confName{CVPR}
\def\confYear{2026}

\title{ Generative Texture Diversification of 3D Pedestrians for Robust Autonomous Driving Perception}

\author{
Arka Bhowmick$^{1}$, Enes \"Ozeren$^{1}$, Ahmed Abdullah$^{2}$, Oliver Wasenm\"uller$^{2}$\\
$^{1}$BIT Technology Solutions GmbH, Germany\\
$^{2}$Mannheim University of Applied Sciences, Germany\\
{\tt\small
{\ arka.bhowmick@bit-ts.de, enozeren@gmail.com}, {\{a.abdullah, o.wasenmueller\}}@th-mannheim.de
}
}

\begin{document}
\maketitle

\input{sec/0_abstract}
\input{sec/1_introduction}
\input{sec/2_Related_Work}
\input{sec/3_Experiments}

\input{sec/results_and_evaluation}
\input{sec/conclusion}

{
    \small
    \bibliographystyle{ieeenat_fullname}
    \bibliography{main}
}


\end{document}

%% file: preamble.tex
\usepackage{pgfplots}
\pgfplotsset{compat=1.18}
\usepackage{tcolorbox}
\usepackage{graphicx}
\usepackage{booktabs}
\usepackage{comment}

%% file: sec/0_abstract.tex
\begin{abstract}
In recent years, autonomous driving has significantly increased the demand for high-quality data to train 2D and 3D perception models for safety-critical scenarios. Real-world datasets struggle to meet this demand as requirements continuously evolve and large-scale annotated data collection remains costly and time-consuming making synthetic data a scalable, practical and controllable alternative.
Pedestrian detection is among the most safety-critical tasks in autonomous driving. In this paper, we propose a simple yet effective method for scaling variability in 3D pedestrian assets for synthetic scene generation. Starting from a single 3D base asset, we generate multiple distinct pedestrian instances by synthesizing diverse facial textures and identity-level appearance variations using StyleGAN2 and automatically mapping them onto 3D meshes. This approach enables scalable appearance-level asset diversification without requiring the design of new geometries for each instance.
Using the assets, we construct synthetic datasets and study the impact of mixing real and synthetic data for RGB-based object detection. Through complementary experiments, we analyze geometry-driven distribution shifts in point cloud perception for 3D object detection. Our findings demonstrate that controlled synthetic diversification improves robustness in 2D detection while revealing the sensitivity of 3D perception models to geometric domain gaps.
Overall, this work highlights how generative AI enables scalable, simulation-ready pedestrian diversification through controlled facial texture synthesis, along with the benefits and limitations of cross-domain training strategies in autonomous driving pipelines.
\end{abstract}

%% file: sec/1_introduction.tex
\section{Introduction}
\label{sec:intro}
Synthetic data has become a critical asset in machine learning, particularly in autonomous driving, where large-scale annotated datasets \cite{cabon2020virtualkitti2, gaidon2016virtualworldsproxymultiobject, Ros_2016_CVPR, scucchia2025gamingresearchgtav} are essential for robust perception. Acquiring real-world driving data is both time-consuming and expensive, often requiring hundreds of kilometers of driving and thousands of hours of recording. Even then, critical scenarios and rare edge cases may remain underrepresented. 
Moreover, privacy constraints and annotation costs further limit the scalability of real-world data collection.
For example, pedestrian instances in real datasets may be insufficient to ensure robust human detection under diverse environmental conditions.
To improve perception robustness, training datasets must exhibit substantial variability. In synthetic environments, scene elements such as buildings, vehicles, traffic infrastructure, and pedestrians can be controlled and diversified. Among these, pedestrian detection remains one of the most safety-critical tasks in autonomous driving. Consequently, generating diverse pedestrian assets efficiently is of high importance. However, manually creating large numbers of realistic 3D human assets is both costly and labor-intensive. Existing approaches typically rely on collections of handcrafted models, which require generating new meshes, re-rigging assets, or performing computationally expensive optimization and simulation, thereby making large-scale deployment time-consuming. Alternatively, some methods primarily introduce diversity through pose and animation variation while keeping the identity appearance largely unchanged, resulting in limited diversity in facial characteristics and fine-grained semantic attributes.

In this paper, we propose a simple yet effective approach to generate multiple distinct instances of pedestrians from a single 3D base model by varying facial textures. We retain a single rigged base model and scale diversity purely through controlled generative texture synthesis in UV space. We train a StyleGAN2 \cite{Karras_2020_CVPR} on FFHQ-UV \cite{bai2023ffhquvnormalizedfacialuvtexture} to generate realistic and diverse facial textures. The generated textures are evaluated to filter-out artifacts and ensure consistency, and are then applied to the 3D base model to create a distinct pedestrian asset. This enables rapid batch-wise diversification of simulation-ready assets, significantly reducing time and computational overhead compared to geometry-based generation methods.

Using these diversified pedestrian assets, we construct synthetic datasets using a rendering pipeline by combining the resulting instances with other scene assets. A mix of synthetic and real datasets is then used to train pedestrian detection models. Experimental results demonstrate improved robustness compared to training solely on real-world datasets such as KITTI \cite{DBLP:journals/corr/abs-2109-13410}, BDD100K \cite{yu2020bdd100k} and A2D2 \cite{geyer2020a2d2}.\newline
The main contributions of this paper are threefold. 
\begin{itemize}
\item First, we propose a novel and scalable method for generating diverse pedestrian instances from a single 3D base model through controlled manipulation of facial textures, enabling efficient variability without redesigning geometry. Our method requires only a single annotated 3D human model and a texture synthesis pipeline, making it practical for large-scale deployment.
\item Second, we conduct a systematic cross-dataset evaluation of both 2D and 3D object detection under controlled real–synthetic data mixing, providing quantitative evidence that domain bias is more pronounced in 3D detection than in 2D detection and offering insights into geometry-driven distribution shifts in point-cloud based detection. This analysis highlights modality-specific challenges in sim-to-real transfer.
\item Third and most importantly, we demonstrate that careful dataset mixing provides a meaningful robustness and performance boost in RGB-based object detection. This highlights the practical benefits of incorporating synthetic data into cross-domain training of autonomous driving systems. Our findings suggest that even limited but diverse synthetic augmentation can significantly improve real-world performance.
\end{itemize}

%% file: sec/2_Related_Work.tex
\section{Related Work}
\textbf{Generative Adversarial Networks.}
Since their introduction, GANs \cite{goodfellow2014generativeadversarialnetworks} have since achieved remarkable success in high quality image synthesis which were further enhanced in style-based architectures like StyleGAN \cite{karras2019stylegan} and StyleGAN2 \cite{Karras_2020_CVPR}, demonstrating control and realism through adaptive instance normalization and modulated convolutions. In particular, StyleGAN2 further developed stability and perceptual quality via path length regularization and weight demodulation, becoming a standard baseline for photorealistic face generation. Although, predominantly used in image synthesis, fewer works investigate their behavior when trained directly on UV texture space, where geometric correspondence is preserved and pose variation is removed.

High-resolution facial datasets such as Flickr-Faces-HQ \cite{karras2019stylegan} have enabled learning-based synthesis approaches, and several works have explored 3D-aware generative models and texture completion in UV coordinates.
However, studies analyzing how GAN training helps in UV texture generation and its variability, remain limited. Our work contributes to this direction by analyzing StyleGAN2 trained on UV face textures and studying its behavior under multiple evaluation metrics.

\textbf{Evaluation of Generative Models.} 
The Fréchet Inception Distance (FID) \cite{FID_DBLP:journals/corr/HeuselRUNKH17} has become a standard metric for measuring distributional similarity between real and generated samples in feature space. Other works \cite{2022role} discovered an inherent bias of the pretrained Inception Model towards ImageNet classes, and proposed using the embedding space of models such as CLIP \cite{clip_radford2021learningtransferablevisualmodels} for evaluating the distribution alignment between generated and real images. Despite widespread adoption, discrepancies between distribution-based metrics (e.g., FID) and semantic metrics remain underexplored, particularly in structured domains such as UV-space textures. Our analysis reveals divergences between semantic recall and classical distributional recall during late-stage training, highlighting potential limitations of standard evaluation protocols.

\textbf{Synthetic Datasets.}
The increasing demand for high-quality training data has led to a growing interest in synthetic datasets, which offer cost-effective and efficient alternatives to real-world data \cite{oehrisynth, abdullahboosting, Song_2024, fang2024data}. A comprehensive survey by \textit{Song et al.} \cite{Song_2024} provides an extensive overview of various synthetic datasets, analyzing their characteristics such as the volume of the data set, the diversity of scene conditions, the types of sensor and the formats of the labels. The study also explores the performance of 2D object detection models trained on synthetic data and tested on real world datasets. However, it does not shed light on cross-domain evaluation and point cloud based perception. \textit{Talwar et al} \cite{talwar_et_al} conducted experiments using YOLOv3 \cite{yolo_v3}, comparing the performance of models trained on synthetic datasets versus real datasets when tested on real-world data. While this research offers valuable insights, it is restricted to 2D object detection on RGB-images and does not include point cloud based perception. The detection tasks were limited to vehicles only and did not include data mixing during training, which we have shown in our work.

%% file: sec/3_Experiments.tex
\section{Method}
\subsection{Generating Face Textures}
\begin{figure}[t]
  \centering
  \includegraphics[width=0.9\linewidth]{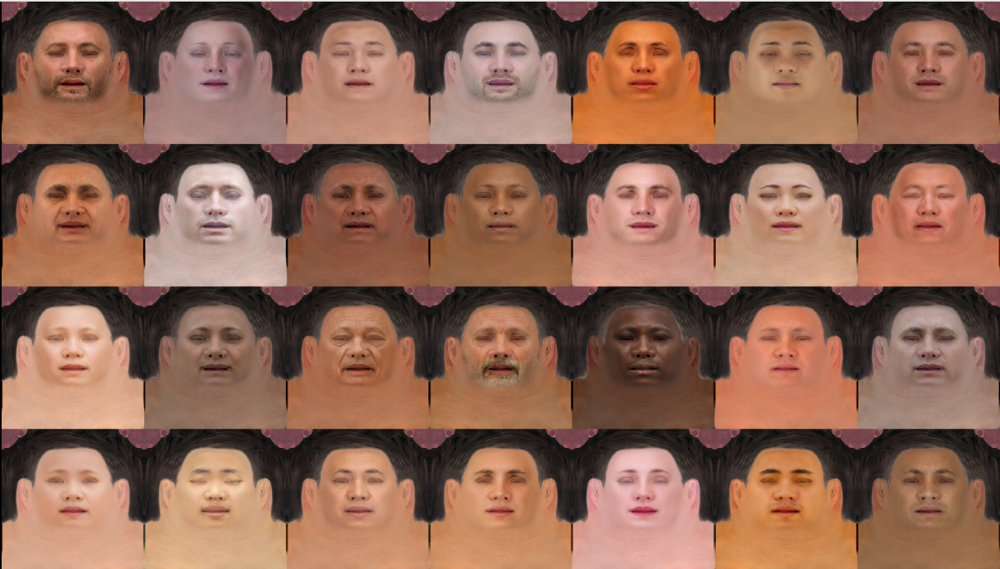}
  \caption{Random samples of unconditionally generated UV face textures from the trained StyleGAN2  \cite{Karras_2020_CVPR} model, illustrating diversity and structural consistency in texture space.}
  \label{fig:random_generated_texture}
\end{figure}

In this section, we elaborate on the process of generating face textures using StyleGAN2 \cite{Karras_2020_CVPR}.
For training the StyleGAN2, we have used the FFHQ-UV dataset \cite{bai2023ffhquvnormalizedfacialuvtexture}. The dataset consists of 50000 face textures that have neutral lighting, neutral expressions and showing cleaned facial areas which are essential for realizing 3D facial models. For the training of the StyleGAN2, 24000 textures were selected and out of which 11K were male textures and 13K female textures. Textures were chosen to cover all age groups. In Table \ref{tab:age_distribution} you may see the number of textures for each age group. This representation of each age group was necessary to make sure that the trained model was able to generate textures for every age group. In Figure \ref{fig:random_generated_texture}, we show unconditionally-generated textures using the trained StyleGAN2 model. 
\begin{table}[htbp]
  \caption{Age distribution of the dataset.}
  \label{tab:age_distribution}
  \centering
  \begin{tabular}{@{}lc@{}}
    \toprule
    Age Group (Years) & Count \\
    \midrule
    $< 10$      & 2991 \\
    $10$--$20$  & 1783 \\
    $21$--$30$  & 6044 \\
    $31$--$40$  & 6645 \\
    $41$--$50$  & 4159 \\
    $51$--$60$  & 2082 \\
    $61$--$70$  & 616  \\
    $71$--$80$  & 35   \\
    \bottomrule
  \end{tabular}
\end{table}

\subsection{Controlling Face Attributes}
In default StyleGAN2 unconditional-generation, facial textures are generated randomly without explicit control over semantic attributes such as age or gender. The resulting attribute distribution follows that of the training dataset. However, for scenario-specific pedestrian synthesis (e.g., school environments) or for enforcing demographic distributions, controlled generation is required. Naively sampling and filtering random outputs is computationally inefficient. Therefore, we introduce a latent space manipulation approach to enable attribute control during generation.

StyleGAN2 generates images from a latent vector $\mathbf{z} \in \mathbb{R}^{512}$ sampled from a random seed. The mapping network (a fully connected neural network) transforms $\mathbf{z}$ into an intermediate latent vector $\mathbf{w} \in \mathbb{R}^{512}$, which encodes disentangled semantic attributes. Our approach manipulates $\mathbf{w}$ directly to control specific attributes in the generated texture.

To learn attribute directions in the latent space, we follow the procedure below:
\begin{itemize}

\item \textbf{Data Generation}: We generate 5,000 facial textures from randomly sampled latent vectors $\mathbf{z}$ and store the corresponding intermediate latent vectors $\mathbf{w}$.

\item \textbf{Annotation}: An external annotator model, that was trained using the annotated FFHQ-UV dataset \cite{bai2023ffhquvnormalizedfacialuvtexture} assigns binary labels to each generated texture across multiple attributes (e.g., male, beardless, old, young, very young). Since the annotator is multi-label, multiple attributes can be active simultaneously.

\item \textbf{Linear Attribute Direction Learning}: For each attribute, a linear Support Vector Machine (SVM) is trained using $\mathbf{w}$ as input and the binary attribute label as output. The normal vector of the learned hyperplane represents the semantic direction of that attribute in the latent space.

Given an attribute direction vector $\mathbf{d}$, attribute manipulation is performed via linear interpolation:

\begin{equation}
\mathbf{w}_{\text{manipulated}} = \mathbf{w} + \alpha \mathbf{d},
\end{equation}

where $\alpha$ controls the strength of the attribute modification. Shifting $\mathbf{w}$ along $\mathbf{d}$ increases or decreases the presence of the corresponding attribute in the generated texture.
This method enables efficient and controllable facial attribute synthesis without retraining the generative model.
\end{itemize}
\begin{figure}[t]
  \centering
  \includegraphics[width=0.9\linewidth]{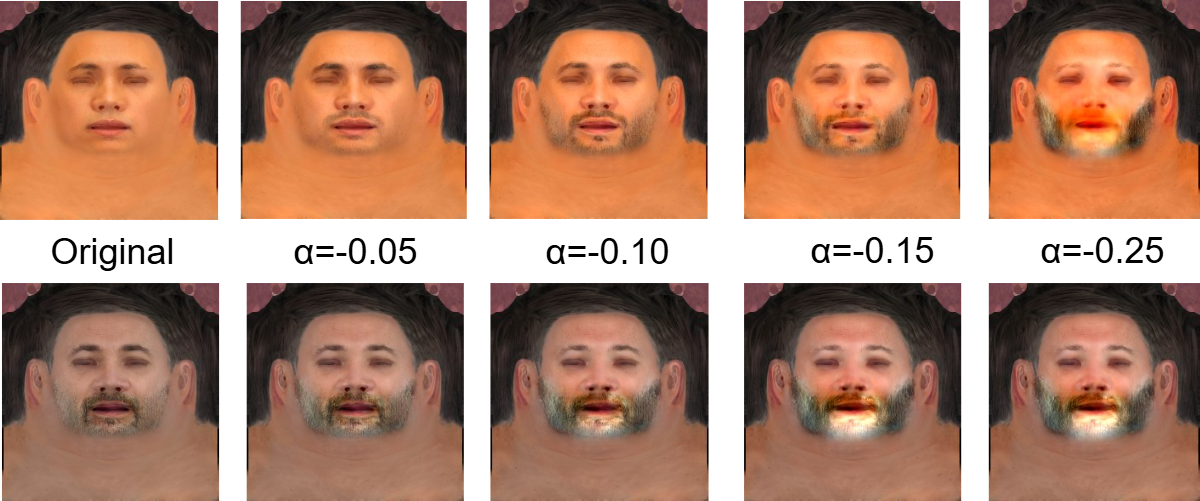}
  \caption{Latent-space manipulation of the beard attribute for two identities by varying the step size $\alpha$. Moderate changes in $\alpha$ produce controlled semantic edits, while larger magnitudes introduce visible distortions and degradation in texture consistency}
  \label{fig:change_attribute_1}
\end{figure}
In Figure \ref{fig:change_attribute_1} we can see that by changing $\alpha$ we can alter the \textit{beard} property but at the same time changing $\alpha$ too much might distort the texture. To ensure stable and controlled attribute manipulation, we introduce two generation-time parameters: the truncation–$\psi$ parameter and a distance-to-boundary parameter.

The truncation–$\psi$ parameter, adopted from StyleGAN2 \cite{Karras_2020_CVPR}, regularizes latent vectors during inference. After computing the mean intermediate latent vector $\mathbf{w_{mean}}$ during training, each latent vector $\mathbf{w}$ is interpolated toward this mean at inference time. The parameter $\psi \in [0,1]$ controls this interpolation, where $\psi=1$ leaves $\mathbf{w}$ unchanged and $\psi=0$ collapses it to $\mathbf{w_{mean}}$. Incorporating truncation into our manipulation framework improves stability and preserves realism during attribute control.

The second parameter is introduced to prevent over-manipulation of attributes. Directly applying a fixed step size may distort textures, particularly when the target attribute is already present. To address this, we adapt the manipulation strength based on the signed distance of the latent vector to the SVM decision boundary. Latent vectors far from the boundary (i.e., strongly lacking the attribute) are modified using a larger step size, while vectors near or beyond the boundary receive smaller adjustments. This adaptive distance-to-boundary scaling prevents the latent vector from drifting outside its natural distribution and avoids visual artifacts.

Together, truncation regularization and distance-aware step size adaptation enable stable, semantically consistent attribute manipulation during facial texture generation. Consequently, the attribute changing process is governed by the equations below:
\begin{equation}
\mathbf{w}_{\text{mean}} 
= \mathbb{E}_{\mathbf{z}} \left[ f(\mathbf{z}) \right]
\label{eq:w_mean}
\end{equation}
\begin{equation}
\mathbf{w}_{\text{manipulated}} 
= \mathbf{w} 
+ \alpha \mathbf{d}
\label{eq:w_manipulated}
\end{equation}
\begin{equation}
\mathbf{w}' 
= \mathbf{w}_{\text{mean}} 
+ \psi \left( 
\mathbf{w}_{\text{manipulated}} 
- \mathbf{w}_{\text{mean}} 
\right)
\label{eq:truncation}
\end{equation}
\begin{figure}[t]
  \centering
  \includegraphics[width=0.8\linewidth]{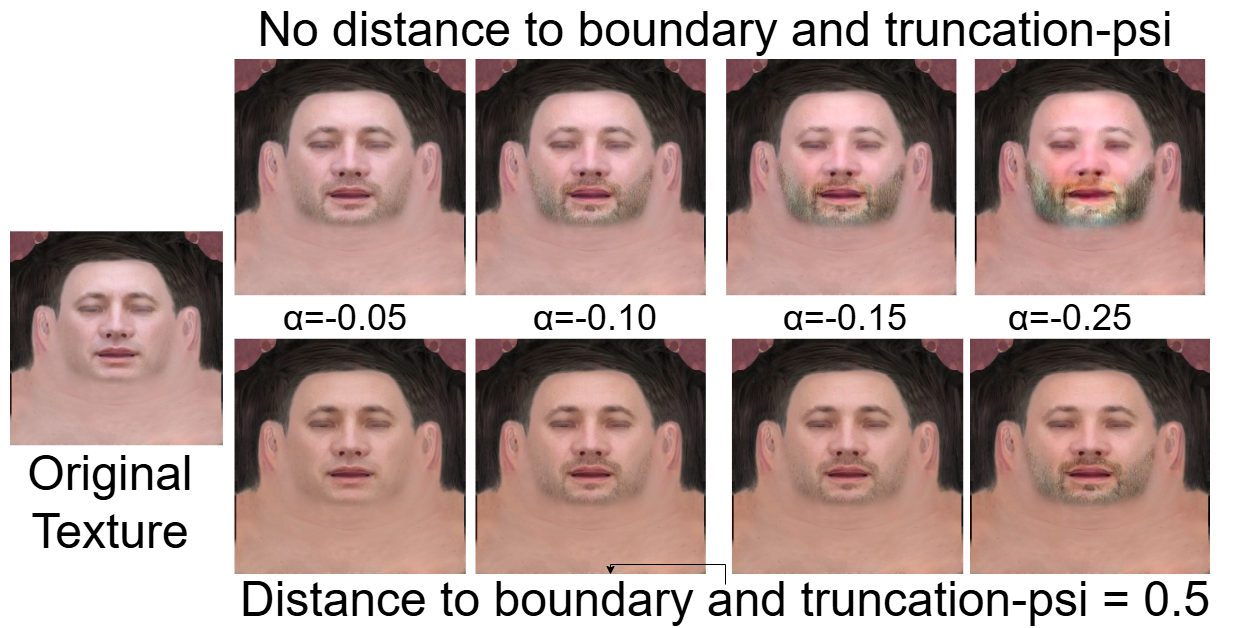}
  \caption{Modifying beard attribute of a face with $\psi$ and distance to boundary parameter showing that the distortion of the texture is controlled.}
  \label{fig:attribute_change_truncation_psi}
\end{figure}

\subsection{Generating Different Instances of The Same 3D Pedestrian Model}
In the final stage of the pipeline, the manipulated facial texture is applied to a high-resolution 3D face mesh generated using the FFHQ-UV framework. The framework provides a UV-consistent facial geometry that aligns with the texture space used during StyleGAN2-based generation. Since both the 3D mesh and the UV-mapped facial textures are derived within the same coordinate and parameterization space, texture projection is performed seamlessly without additional alignment. This guarantees geometric consistency and prevents distortions during rendering.

The resulting textured 3D face models are evaluated to verify semantic correctness, visual fidelity, and attribute consistency. Only validated instances are propagated to the next stage.

\begin{figure}[t]
  \centering
  \includegraphics[width=0.8\linewidth]{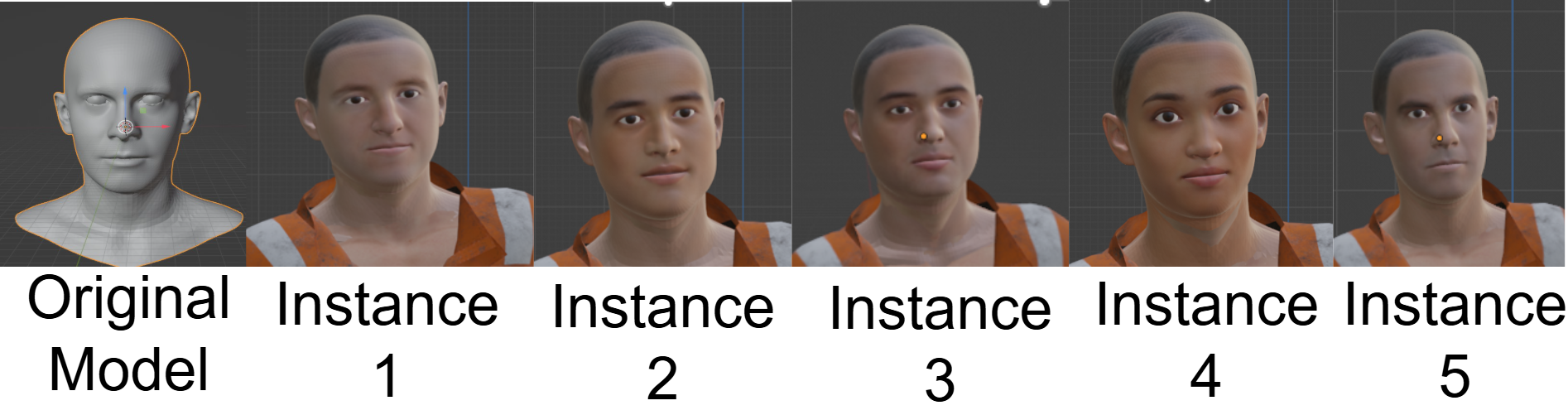}
  \caption{Generated 5 different pedestrian instances from the same 3D asset.}
  \label{fig:same_model_diff_instance}
\end{figure}

Subsequently, the textured facial mesh is integrated into an existing full-body pedestrian model from an internal asset database. The integration process replaces the original head geometry or texture while preserving skeletal structure, rigging, and animation compatibility. This modular design allows multiple distinct facial textures to be applied to the same base pedestrian mesh, generating diverse pedestrian instances without duplicating full-body geometry as shown in Figure \ref{fig:same_model_diff_instance}.

By disentangling face synthesis from body modeling and leveraging UV-consistent asset pipelines, the proposed approach enables scalable pedestrian generation with controlled semantic variation while maintaining compatibility with existing simulation assets.

\subsection{Generating Synthetic Dataset}
\begin{figure}[t]
  \centering
  \includegraphics[width=\linewidth]{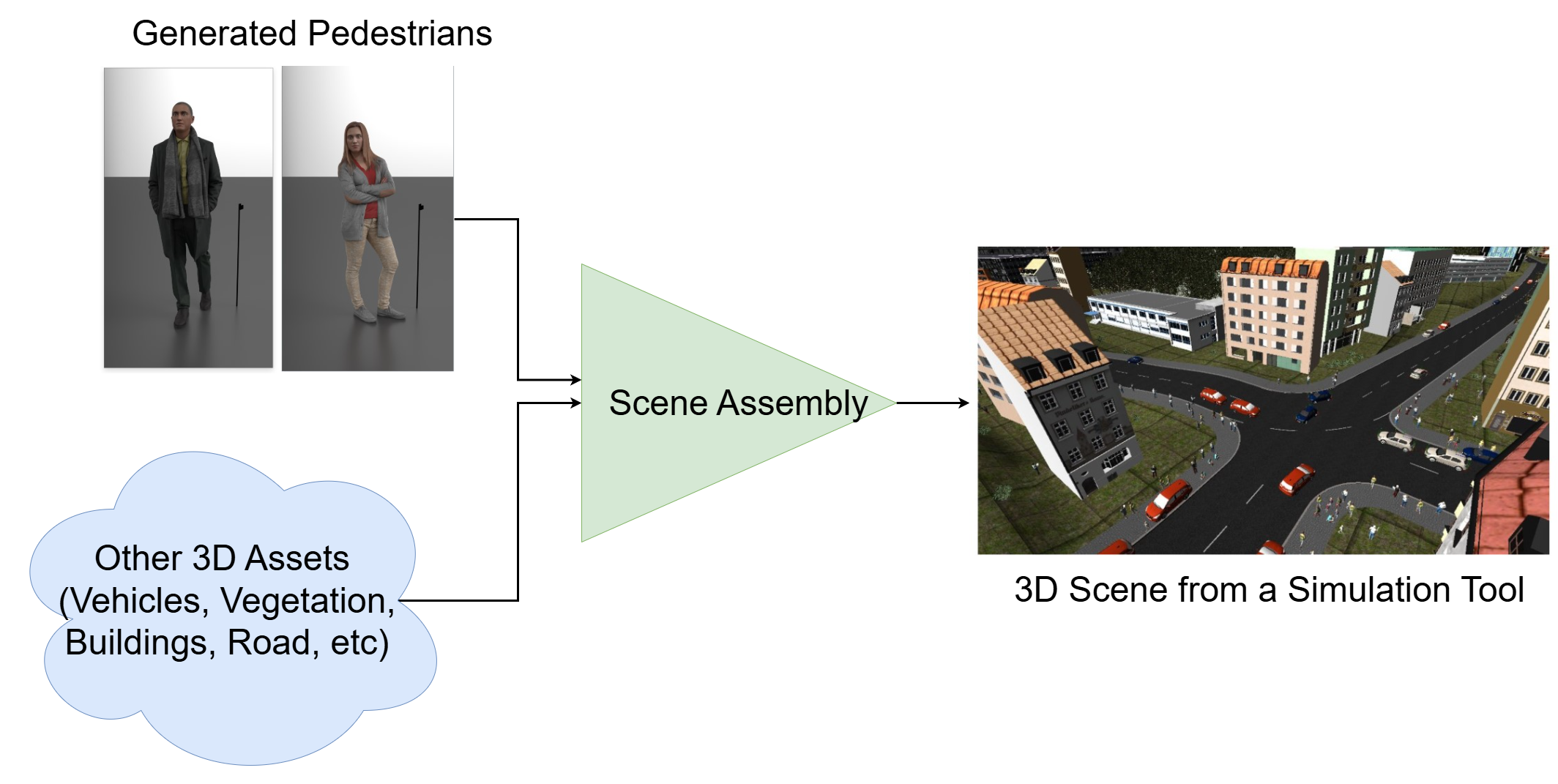}
  \caption{Synthetic data generation snapshot where we combine different static and dynamic assets to create a scenario.}
  \label{fig:scene_generation}
\end{figure}

After the pedestrian asset synthesis, we generate simulation scenarios for autonomous driving validation (Figure \ref{fig:scene_generation}). A scenario consists of a rule based 3D environment with static and dynamic assets including the pedestrians that were generated using the above described method.

%% file: sec/results_and_evaluation.tex
\section{Results and Evaluation}

\subsection{Evaluating and Filtering Generated Textures}
\begin{figure}[t]
  \centering
  \includegraphics[width=\linewidth]{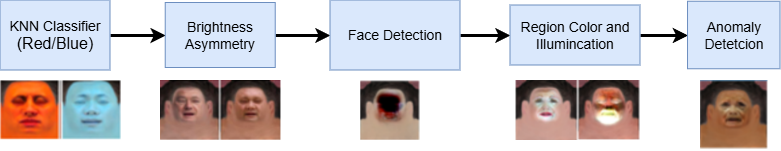}
  \caption{The steps that have been followed to validate individual texture samples.}
  \label{fig:text_eval_1}
\end{figure}

To comprehensively evaluate the realism of the generated textures, dataset-level metrics such as FID \cite{FID_DBLP:journals/corr/HeuselRUNKH17}, Precision, and Recall \cite{precisionrecallkynkaanniemi2019improved} were used. Both Inception \cite{Szegedy2016InceptionV3} and CLIP \cite{clip_radford2021learningtransferablevisualmodels} were utilized as feature extractors. In Table \ref{tab:model_comparison} we show a comparison among 3 different StyleGAN2 \cite{Karras_2020_CVPR} checkpoints.
\begin{table}[t]
\centering
\small
\caption{Mean and standard deviation of 3 StyleGAN2 \cite{Karras_2020_CVPR} checkpoints in terms of FID \cite{FID_DBLP:journals/corr/HeuselRUNKH17}, Precision, and Recall \cite{precisionrecallkynkaanniemi2019improved} scores. '-CLIP' indicates that CLIP \cite{clip_radford2021learningtransferablevisualmodels} is used for feature extraction. Lower FID with higher Precision and Recall indicates better model.}
\label{tab:model_comparison}
\begin{tabular}{lccc}
\toprule
\textbf{Metric} & \textbf{Ckpt-1} & \textbf{Ckpt-2} & \textbf{Ckpt-3} \\
\midrule
FID              & 8.14$\pm$0.02 & 9.04$\pm$0.02 & 8.30$\pm$0.03 \\
FID-CLIP         & 0.03$\pm$0.00 & 1.69$\pm$0.01 & 1.48$\pm$0.00 \\
Precision        & 0.64$\pm$0.00 & 0.66$\pm$0.00 & 0.63$\pm$0.00 \\
Precision-CLIP   & 0.63$\pm$0.00 & 0.84$\pm$0.00 & 0.80$\pm$0.00 \\
Recall           & 0.03$\pm$0.00 & 0.03$\pm$0.00 & 0.05$\pm$0.00 \\
Recall-CLIP      & 0.46$\pm$0.00 & 0.01$\pm$0.00 & 0.02$\pm$0.00 \\
\bottomrule
\end{tabular}
\end{table}
\newline At the individual sample level we use multi-stage sample validation to examine the generated textures for undesirable artifacts and defects such as color-tint errors, brightness asymmetries, and regional color deviations (Figure \ref{fig:text_eval_1}). For color-tint errors, we fit a KNN classifier on a small dataset (~80 samples) curated from the trained StyleGAN2, where samples are classified as either normal, blue-tint or red-tint. Next, we use the brightness symmetry error from FFHQ-UV \cite{bai2023ffhquvnormalizedfacialuvtexture} to filter-out samples where one side of the texture is dimmer or brighter than the other. For region-color deviations, two functions are utilized. The first function evaluates illumination consistency across predefined facial regions by converting them to YUV space, extracting the luminance (Y) channel, applying Gaussian blur, and computing pairwise L1 differences to ensure uniform lighting. The second function assesses color consistency by comparing blurred RGB regions of the face against the neck region using L1 distance, verifying that overall color variation remains below a predefined threshold. Finally we use a ResNet50 \cite{he2016resnet} trained on FFHQ-UV \cite{bai2023ffhquvnormalizedfacialuvtexture}  for anomaly detection of individual textures. 

\subsection{Texture Mapping Evaluation}
\begin{figure}[t]
  \centering
  \includegraphics[width=0.75\linewidth]{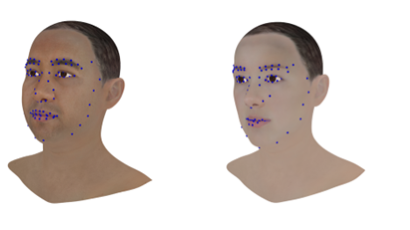}
  \caption{2D texture is mapped onto the 3D model, rendered and facial landmarks are overlayed for texture mapping analysis. }
  \label{fig:mapping}
\end{figure}
After mapping a generated texture onto a 3D face model, we evaluate distributional consistency using 3D-FID and 3D-KID (Kernel Inception Distance). Specifically, a Blender-based pipeline is used to render the 3D base head mesh from a random viewpoint, after mapping a texture image on the 3D head model (Figure \ref{fig:mapping}). This process is repeated to construct a dataset of 50k rendered photos from 50k generated textures. For creating a reference dataset, we follow the same process but use FFHQ-UV \cite{bai2023ffhquvnormalizedfacialuvtexture} to construct facial textures from samples taken from the Chicago Face Dataset \cite{ma2015chicago}. Low 3D-FID and 3D-KID values (Table \ref{tab:3d_fid_kid}) indicate strong global similarity between mapped and reference textures, reflecting high realism and distributional fidelity in the rendered 3D space.
\begin{table}[t]
\centering
\small
\caption{Evaluation of mapped textures using 3D-FID and 3D-KID. Lower values indicate better distributional similarity to the reference dataset.}
\label{tab:3d_fid_kid}
\begin{tabular}{lcc}
\toprule
\textbf{Metric} & \textbf{3D-FID} & \textbf{3D-KID} \\
\midrule
Mapped Textures vs.\ Reference & 3.597 & 0.013$\pm$0.001 \\
\bottomrule
\end{tabular}
\end{table}

\subsection{Evaluating Impact of Synthetic Data}
\subsubsection{Datasets}
For our experiments we have used both real datasets and synthetic dataset. As a part of the real dataset we have used KITTI \cite{DBLP:journals/corr/abs-2109-13410}, BDD100K \cite{yu2020bdd100k} and A2D2 \cite{geyer2020a2d2} datasets. With respect to synthetic data we have used internally generated synthetic datasets.
For 2D object detection tasks we have used the KITTI, BDD100K and internally generated  synthetic dataset. For 3D object detection task we utilize the LiDAR point cloud data from KITTI, A2D2 and internally generated synthetic dataset. As BDD100K does not contain LiDAR data, it was replaced with A2D2.
\subsubsection{Object (Pedestrian) Detection Models}
For 2D pedestrian detection task the YOLOv7 \cite{Wang_2023_CVPR} model was used and for 3D pedestrian detection task SECOND \cite{openpcdet2020} model was used. These models were selected for two primary reasons: 
\begin{itemize}
    \item They are well-established algorithms with a proven track record of success in object detection.
    \item  Their open-source repositories facilitate ease of implementation.
\end{itemize}

\begin{figure}[t]
  \centering
  \includegraphics[width=0.75\linewidth]{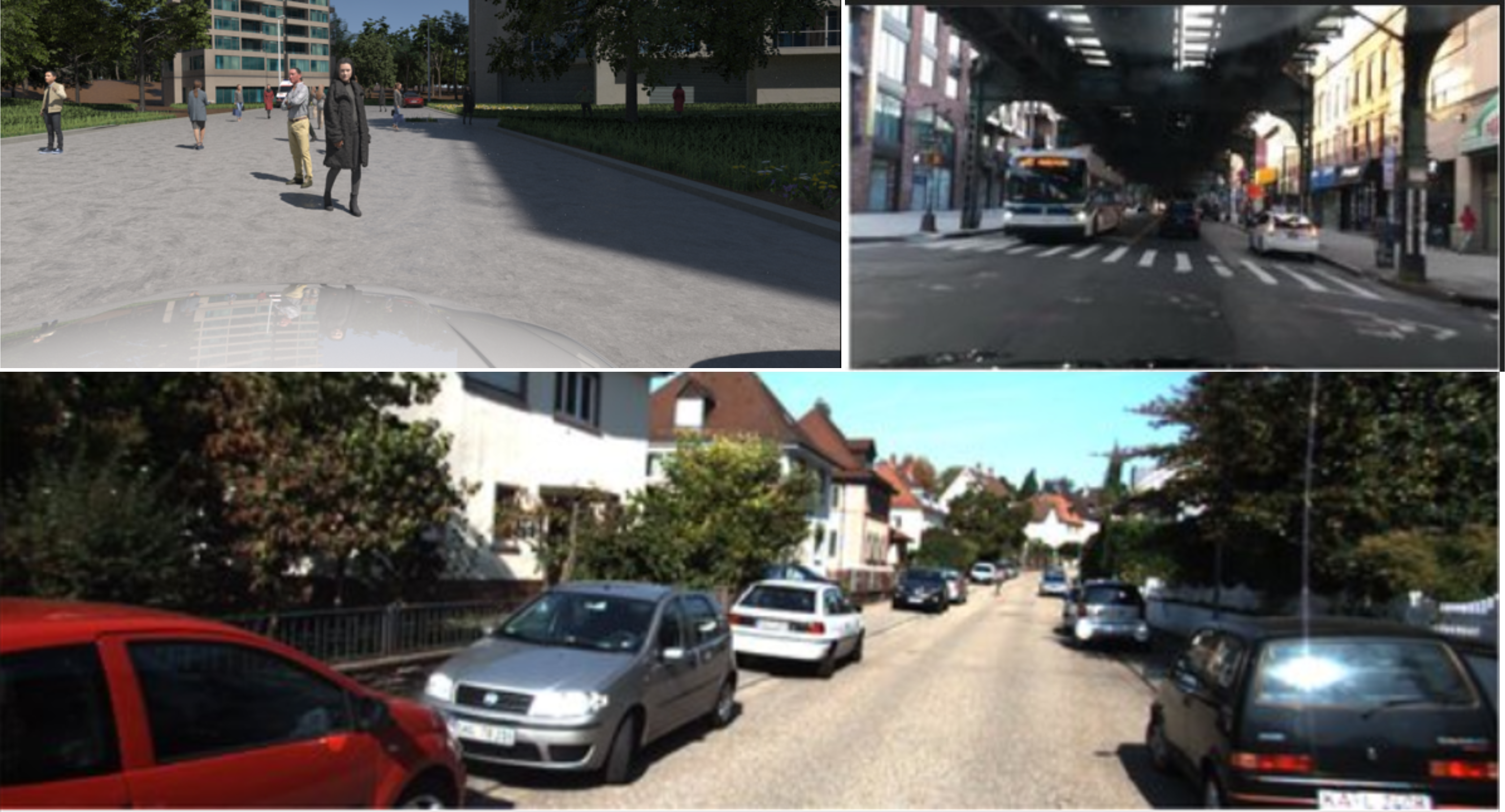}
  \caption{Image examples from three datasets: Synthetic data (top left), BDD100K (top right), and
KITTI (bottom).}
  \label{fig:image_samples_v2}
\end{figure}

\begin{figure}[t]
  \centering
  \includegraphics[width=0.75\linewidth]{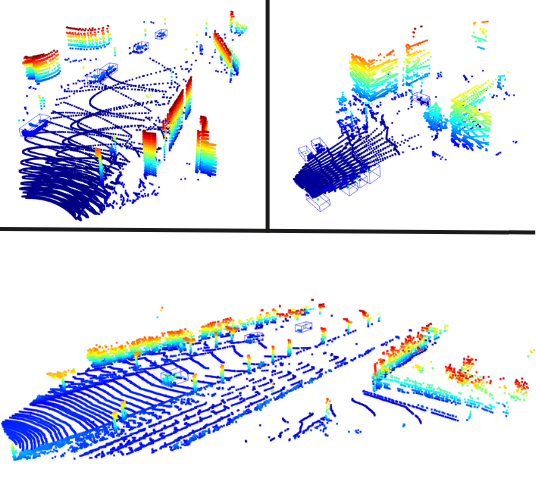}
  \caption{Point cloud examples from three datasets: Synthetic Point Cloud (top left), A2D2 (top right), and
KITTI (bottom).}
  \label{fig:point_cloud_samples_v2}
\end{figure}

\subsubsection{2D Object (Pedestrian) Detection}
For the 2D object detection task we trained $7$ models, each using a different dataset while keeping the hyperparameters consistent across all models. Table \ref{tab:2d_data_composition} presents the volumes of the training datasets used for each of the 7 models. For training, we initialize the model parameters randomly and avoid transfer learning to maintain a controlled experimental environment that allows us to assess the impact of the different datasets used. The training process employs the Stochastic Gradient Descent algorithm, with an initial learning rate of $0.007$ and a learning rate scaling factor of $0.1$. A weight decay of $0.0005$ is applied, and the default data augmentations from the YOLOv7 \cite{Wang_2023_CVPR} codebase. The batch size is set to $4$ per GPU(16 in total), and the model is trained for $50$ epochs with $4$ Nvidia RTX 6000 GPUs. All 7 models are trained with the same hyperparameters and we observe that $50$ epochs are sufficient for all models to converge with respect to the validation loss.

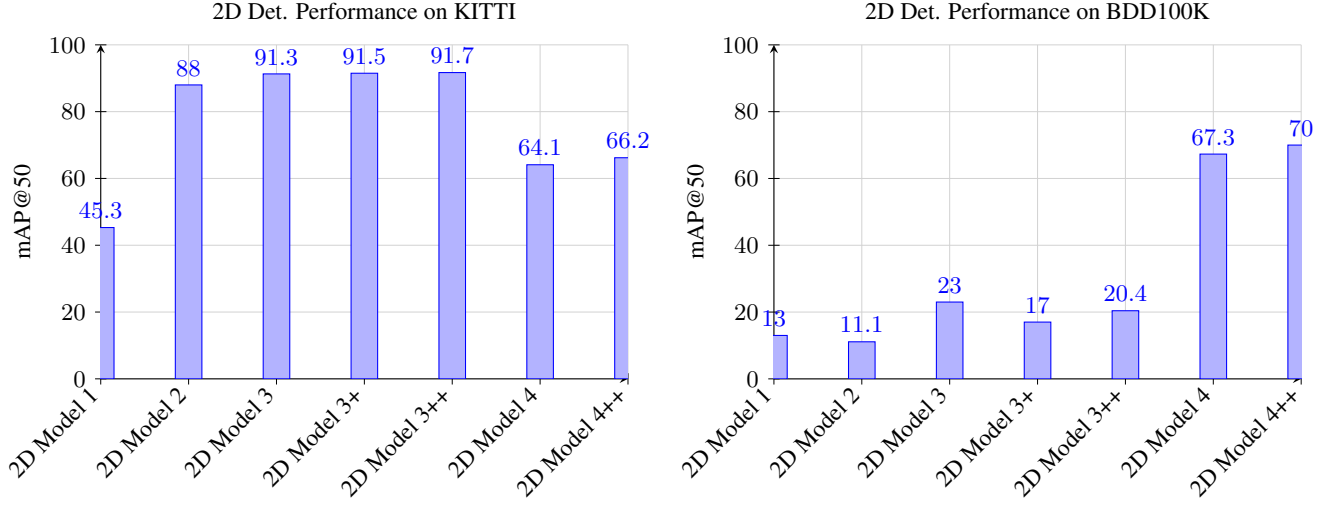
\begin{figure*}[t]
\centering

\begin{minipage}{0.49\textwidth}
\centering
\begin{tikzpicture}
\begin{axis}[
    ybar,
    bar width=10pt,
    width=\linewidth,
    height=6.0cm,
    ymin=0, ymax=100,
    ylabel={mAP@50},
    title={2D Det. Performance on KITTI},
    symbolic x coords={
        2D Model 1,
        2D Model 2,
        2D Model 3,
        2D Model 3+,
        2D Model 3++,
        2D Model 4,
        2D Model 4++
    },
    xtick=data,
    x tick label style={rotate=45, anchor=east, font=\small},
    yticklabel style={font=\small},
    label style={font=\small},
    title style={font=\small},
    nodes near coords,
    nodes near coords style={font=\small},
    nodes near coords align={vertical},
    enlarge x limits=0.15,
    axis lines=left,
    tick style={black},
    grid=both,
    grid style={line width=.1pt, draw=gray!25},
    major grid style={line width=.2pt, draw=gray!35}
]
\addplot coordinates {
    (2D Model 1,45.3)
    (2D Model 2,88.0)
    (2D Model 3,91.3)
    (2D Model 3+,91.5)
    (2D Model 3++,91.7)
    (2D Model 4,64.1)
    (2D Model 4++,66.2)
};
\end{axis}
\end{tikzpicture}
\end{minipage}
\hfill
\begin{minipage}{0.49\textwidth}
\centering
\begin{tikzpicture}
\begin{axis}[
    ybar,
    bar width=10pt,
    width=\linewidth,
    height=6.0cm,
    ymin=0, ymax=100,
    ylabel={mAP@50},
    title={2D Det. Performance on BDD100K},
    symbolic x coords={
        2D Model 1,
        2D Model 2,
        2D Model 3,
        2D Model 3+,
        2D Model 3++,
        2D Model 4,
        2D Model 4++
    },
    xtick=data,
    x tick label style={rotate=45, anchor=east, font=\small},
    yticklabel style={font=\small},
    label style={font=\small},
    title style={font=\small},
    nodes near coords,
    nodes near coords style={font=\small},
    nodes near coords align={vertical},
    enlarge x limits=0.15,
    axis lines=left,
    tick style={black},
    grid=both,
    grid style={line width=.1pt, draw=gray!25},
    major grid style={line width=.2pt, draw=gray!35}
]
\addplot coordinates {
    (2D Model 1,13.0)
    (2D Model 2,11.1)
    (2D Model 3,23.0)
    (2D Model 3+,17.0)
    (2D Model 3++,20.4)
    (2D Model 4,67.3)
    (2D Model 4++,70.0)
};
\end{axis}
\end{tikzpicture}
\end{minipage}
\caption{Test-set 2D object detection performance (mAP@50) for YOLOv7 trained on different dataset compositions and evaluated on (left) KITTI and (right) BDD100K.}
\label{fig:2d_dual_eval}
\end{figure*}

\begin{table*}[!t]
\centering
\small
\caption{Training data composition for different 2D object detection models.}
\label{tab:2d_data_composition}
\begin{tabular}{lccccc}
\toprule
\textbf{Model Name} & \textbf{Dataset} & \textbf{Synthetic Frames} & \textbf{KITTI Frames} & \textbf{BDD100K Frames} & \textbf{Total Frames} \\
\midrule
2D Model 1   & Synthetic                    & 6K & 0  & 0  & 6K  \\
2D Model 2   & KITTI                     & 0  & 6K & 0  & 6K  \\
2D Model 3   & Synthetic + KITTI (MIX)      & 2K & 6K & 0  & 8K  \\
2D Model 3+  & Synthetic + KITTI (MIX+)     & 4K & 6K & 0  & 10K \\
2D Model 3++ & Synthetic + KITTI (MIX++)    & 6K & 6K & 0  & 12K \\
2D Model 4   & BDD100K                   & 0  & 0  & 6K & 6K  \\
2D Model 4++ & Synthetic + BDD100K (MIX++)  & 6K & 0  & 6K & 12K \\
\bottomrule
\end{tabular}
\end{table*}
We evaluate cross-dataset generalization by training YOLOv7 under multiple dataset compositions and testing separately on KITTI and BDD100K test sets.
When evaluated on KITTI as shown in Figure \ref{fig:2d_dual_eval} (left), the KITTI-only baseline achieves 88.0\% mAP@50. Incorporating synthetic data progressively improves performance to 91.7\% mAP (3.7\% absolute, 4.2\% relative improvement). This consistent monotonic increase across \textit{MIX}, \textit{MIX+}, and \textit{MIX++} settings suggests that the generated synthetic data introduces complementary variability rather than noise, acting as an effective regularizer. In contrast, models trained solely on BDD100K achieve only 64.1\% mAP on KITTI, indicating substantial domain shift between KITTI and BDD100K.

A reciprocal trend is observed when testing on BDD100K as evident from Figure \ref{fig:2d_dual_eval} (right). The BDD100K only baseline achieves 67.3 mAP, which increases to 70.0 mAP when synthetic data is added (+2.7\% absolute, +4.0\% relative). However, KITTI only training yields 11.1\% mAP on BDD100K, reflecting severe cross-domain degradation (-56.2\% absolute compared to in-domain performance).
Overall, cross-dataset transfer results in large performance drops (up to -76.9\% mAP in extreme cases), quantitatively confirming strong dataset-specific bias. Nevertheless, when synthetic data is combined with target-domain data, it consistently yields statistically meaningful improvements (approx. 3\% to 4\% mAP), demonstrating that dataset diversity enhances robustness even in fully supervised settings.

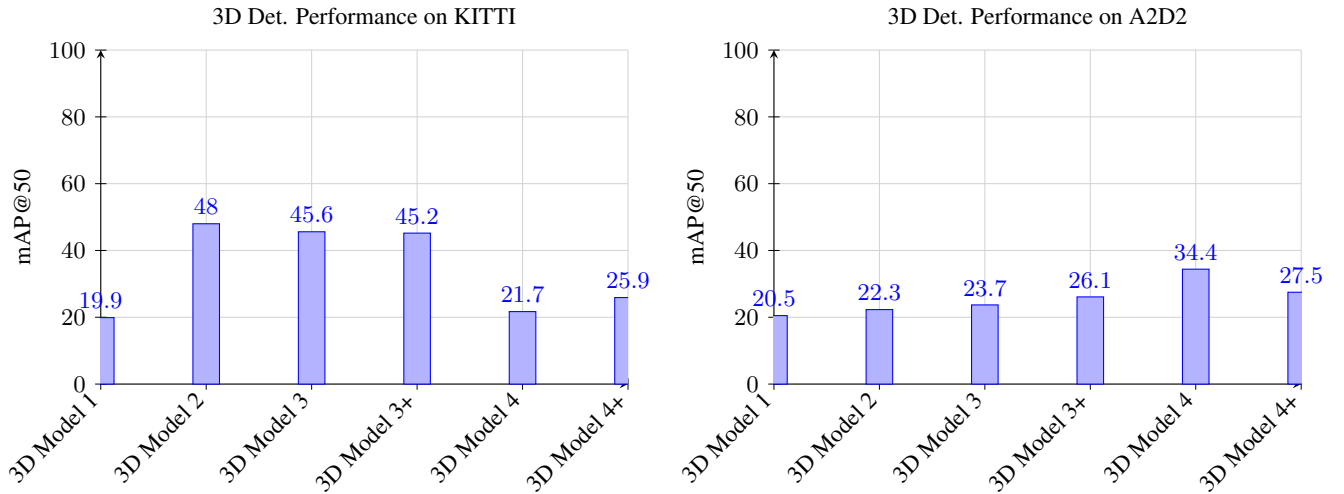
\begin{figure*}[!t]
\centering
\begin{minipage}{0.49\textwidth}
\centering
\begin{tikzpicture}
\begin{axis}[
    ybar,
    bar width=10pt,
    width=\linewidth,
    height=6.0cm,
    ymin=0, ymax=100,
    ylabel={mAP@50},
    title={3D Det. Performance on KITTI},
    symbolic x coords={
        3D Model 1,
        3D Model 2,
        3D Model 3,
        3D Model 3+,
        3D Model 4,
        3D Model 4+
    },
    xtick=data,
    x tick label style={rotate=45, anchor=east, font=\small},
    yticklabel style={font=\small},
    label style={font=\small},
    title style={font=\small},
    nodes near coords,
    nodes near coords style={font=\small},
    nodes near coords align={vertical},
    enlarge x limits=0.15,
    axis lines=left,
    tick style={black},
    grid=both,
    grid style={line width=.1pt, draw=gray!25},
    major grid style={line width=.2pt, draw=gray!35}
]
\addplot coordinates {
    (3D Model 1,19.9)
    (3D Model 2,48.0)
    (3D Model 3,45.6)
    (3D Model 3+,45.2)
    (3D Model 4,21.7)
    (3D Model 4+,25.9)
};
\end{axis}
\end{tikzpicture}
\end{minipage}
\hfill
\begin{minipage}{0.49\textwidth}
\centering
\begin{tikzpicture}
\begin{axis}[
    ybar,
    bar width=10pt,
    width=\linewidth,
    height=6.0cm,
    ymin=0, ymax=100,
    ylabel={mAP@50},
    title={3D Det. Performance on A2D2},
    symbolic x coords={
        3D Model 1,
        3D Model 2,
        3D Model 3,
        3D Model 3+,
        3D Model 4,
        3D Model 4+
    },
    xtick=data,
    x tick label style={rotate=45, anchor=east, font=\small},
    yticklabel style={font=\small},
    label style={font=\small},
    title style={font=\small},
    nodes near coords,
    nodes near coords style={font=\small},
    nodes near coords align={vertical},
    enlarge x limits=0.15,
    axis lines=left,
    tick style={black},
    grid=both,
    grid style={line width=.1pt, draw=gray!25},
    major grid style={line width=.2pt, draw=gray!35}
]
\addplot coordinates {
    (3D Model 1,20.5)
    (3D Model 2,22.3)
    (3D Model 3,23.7)
    (3D Model 3+,26.1)
    (3D Model 4,34.4)
    (3D Model 4+,27.5)
};
\end{axis}
\end{tikzpicture}
\end{minipage}
\caption{Test-set 3D object detection mAP@50 values for SECOND-3D Detection Models trained on different dataset compositions and evaluated on (left) KITTI and (right) A2D2.}
\label{fig:3d_dual_eval}
\end{figure*}
\begin{table*}[!t]
\centering
\small
\caption{Training data composition for different 3D detection models.}
\label{tab:3d_data_composition}
\begin{tabular}{lccccc}
\toprule
\textbf{Model Name} & \textbf{Dataset} & \textbf{Synthetic Frames} & \textbf{KITTI Frames} & \textbf{A2D2 Frames} & \textbf{Total Frames} \\
\midrule
3D Model 1   & Synthetic                    & 4K & 0  & 0  & 4K \\
3D Model 2   & KITTI                      & 0  & 4K & 0  & 4K \\
3D Model 3   & Synthetic + KITTI (MIX)       & 2K & 4K & 0  & 6K \\
3D Model 3+  & Synthetic + KITTI (MIX+)      & 4K & 4K & 0  & 8K \\
3D Model 4   & A2D2                       & 0  & 0  & 4K & 4K \\
3D Model 4+  & Synthetic + A2D2 (MIX+)       & 4K & 0  & 4K & 8K \\
\bottomrule
\end{tabular}
\end{table*}
\subsubsection{3D Object (Pedestrian) Detection}
For the 3D object detection task, we trained $6$ models, each using a different dataset while keeping the hyperparameters consistent across all models. Table \ref{tab:3d_data_composition} presents the volumes of the training datasets used for each of the 6 models. For training the 3D object detection model SECOND \cite{openpcdet2020} we use the same strategy as in 2D object detection and initialize the model weights randomly. We use the AdamW optimizer with initial learning rate $0.001$ and a learning rate scheduler "StepLR" in pytorch with step size $2$ and gamma $0.85$. We also use weight decay with $0.0001$ and default data augmentation from the OpenPCDet codebase \cite{openpcdet2020}. The batch size is $16$ per GPU (64 in total) and we train all of $6$ models for $50$ epochs with $4$ Nvidia RTX 6000 GPUs. All models converged in $50$ epochs.\newline
When evaluated on KITTI as seen in Figure \ref{fig:3d_dual_eval} (left), the KITTI-only model achieves 48.0\% mAP@50. In contrast to the 2D experiments, incorporating synthetic data samples results in a slight performance degradation (45.6\% and 45.2\% mAP), suggesting that heterogeneous point cloud distributions negatively impact 3D feature learning. Models trained solely on synthetic data or A2D2 exhibit severe performance drops on KITTI (19.9\% and 21.7\% mAP), indicating substantial geometric domain shift.
A similar trend is observed when evaluating on A2D2 dataset as seen in Figure \ref{fig:3d_dual_eval} (right). The A2D2-only baseline achieves 34.4\% mAP, while including synthetic data reduces performance to 27.5\% mAP (-6.9\% absolute). Although mixing KITTI and synthetic data slightly improves over single cross-domain training (26.1\% mAP vs. 22.3\% mAP), it remains significantly below in-domain performance. \newline Overall, 3D detection demonstrates markedly stronger dataset bias compared to 2D detection, highlighting the sensitivity of voxel-based models to sensor geometry, point density statistics, and coordinate priors.

\subsubsection{Key Takeaways}
Dataset bias is substantially stronger in 3D detection than in 2D detection.
While mixing datasets improves 2D performance by up to +4\% relative mAP, the same strategy degrades 3D performance by nearly 20\% relative in some settings. 
This contrast highlights that RGB-based models benefit from visual diversity, whereas voxel-based 3D models are highly sensitive to geometric distribution shifts (sensor calibration, point density, coordinate priors). 
These findings emphasize the need for domain-aware normalization or adaptation strategies for multi-dataset 3D training.
\subsubsection{Ablation Study}
We conduct ablation studies for both 2D and 3D pedestrian detection to evaluate the effect of incorporating synthetic data across datasets. As shown in Tables \ref{tab:ablation_synth_vs_real} and \ref{tab:3d_ablation_synth_vs_real}, real-only training yields strong in-domain performance but exhibits limited cross-dataset generalization. In contrast, mixed synthetic-real training consistently improves cross-domain performance for 2D detection (e.g., BDD→KITTI: 49.0 → 50.8), indicating enhanced robustness due to increased appearance diversity. For 3D detection, however, the benefits are less consistent, with performance often degrading under mixed training, highlighting sensitivity to geometric domain shifts between datasets. Overall, these results demonstrate that synthetic data is effective for improving visual robustness in 2D perception, while careful consideration is required when integrating synthetic data for geometry-driven 3D tasks.
\begin{table}[!t]
\centering
\small
\caption{Ablation study for 2D object detetction comparing real-only and mixed synthetic-real training across datasets. All models are evaluated using mAP@50 for pedestrian detection.}
\label{tab:ablation_synth_vs_real}
\resizebox{\linewidth}{!}{
\begin{tabular}{lcccc}
\toprule
\textbf{Training Data} & \textbf{\#Images} & \textbf{BDD100K(map@50)} & \textbf{KITTI(map@50)} & \textbf{Setting} \\
\midrule
BDD100K (real only) & 8K & 68.7 & 49.0 & In-domain \\
Synthetic + BDD100K (50/50) & 8K & 54.9 & 50.8 & Mixed \\
KITTI (real only) & 6K & 11.1 & 88.0 & In-domain \\
Synthetic + KITTI (50/50) & 6K & 18.4 & 45.1 & Mixed \\
\bottomrule
\end{tabular}
}
\end{table}
\vspace{-2mm}
\begin{table}[!t]
\centering
\small
\caption{Ablation study for 3D object detetction comparing real-only and mixed synthetic-real training across datasets. All models are evaluated using mAP@50 for pedestrian detection.}
\label{tab:3d_ablation_synth_vs_real}
\resizebox{\linewidth}{!}{
\begin{tabular}{lcccc}
\toprule
\textbf{Training Data} & \textbf{\#Images} & \textbf{A2D2(map@50)} & \textbf{KITTI(map@50)} & \textbf{Setting} \\
\midrule
A2D2 (real only) & 8K & 27.95 & 21.37 & In-domain \\
Synthetic + A2D2 (50/50) & 8K & 27.50 & 25.90 & Mixed \\
KITTI (real only) & 6K & 11.10 & 12.97 & In-domain \\
Synthetic + KITTI (50/50) & 6K & 25.12 & 26.78 & Mixed \\
\bottomrule
\end{tabular}
}
\end{table}

%% file: sec/conclusion.tex
\section{Conclusion}
In this work, we present an efficient method for appearance-driven identity diversification of 3D pedestrian models through automated generation and controlled manipulation of facial textures. By synthesizing multiple realistic texture variations from a single base model, we enable scalable identity diversity for synthetic scene generation without redesigning 3D geometries.
Through systematic experiments combining real and synthetic data, we demonstrate that diversified synthetic pedestrians improve perception robustness in RGB-based object detection compared to training solely on real-world datasets such as KITTI and BDD100K. Our ablation study further demonstrates that these gains stem from increased data diversity rather than dataset size alone. However, we observe a trade-off between improved generalization and slight reductions in in-domain performance, highlighting the importance of balanced data composition.
Furthermore, our analysis reveals that 3D object detection models exhibit high sensitivity to geometric domain shift. Unlike RGB-based models, point cloud perception can degrade when heterogeneous datasets are mixed, due to differences in sensor geometry, point density, and coordinate distributions.
In general, our findings highlight both the benefits of incorporating synthetic data for improving visual robustness and the limitations of naive multi-dataset training in geometry-driven 3D perception. Due to texture dataset availability, our synthetic augmentation is limited to the facial region of 3D pedestrian assets. Nonetheless, we hope our work motivates future methods to explore the impact of full-asset texture diversification for autonomous perception.

 \section*{Acknowledgments}
This work was partially funded by the German Federal Ministry for Education and Research (BMBF) under the grant KIEDAAS (01IS22037). 

%% file: main.bib
@String(CVPR= {IEEE Conf. Comput. Vis. Pattern Recog.})

@String(ICPR = {Int. Conf. Pattern Recog.})

@String(ACCV  = {ACCV})

@String(ICLR = {Int. Conf. Learn. Represent.})

@String(CVPR  = {CVPR})

@String(ICPR  = {ICPR})

@String(ICLR  = {ICLR})

@InProceedings{Karras_2020_CVPR,
    author = {Karras, Tero and Laine, Samuli and Aittala, Miika and Hellsten, Janne and Lehtinen, Jaakko and Aila, Timo},
    title = {Analyzing and Improving the Image Quality of StyleGAN},
    booktitle = {Proceedings of the IEEE/CVF Conference on Computer Vision and Pattern Recognition (CVPR)},
    month = {June},
    year = {2020}
}

@article{DBLP:journals/corr/abs-2109-13410,
  author       = {Yiyi Liao and
                  Jun Xie and
                  Andreas Geiger},
  title        = {{KITTI-360:} {A} Novel Dataset and Benchmarks for Urban Scene Understanding
                  in 2D and 3D},
  journal      = {CoRR},
  volume       = {abs/2109.13410},
  year         = {2021}
}

@inproceedings{yu2020bdd100k,
  title     = {BDD100K: A Diverse Driving Dataset for Heterogeneous Multitask Learning},
  author    = {Yu, Fisher and Chen, Haofeng and Wang, Xin and Xian, Wenqi and Chen, Yingying and Liu, Fangchen and Madhavan, Vashisht and Darrell, Trevor},
  booktitle = {Proceedings of the IEEE/CVF Conference on Computer Vision and Pattern Recognition (CVPR)},
  year      = {2020}
}

@inproceedings{geyer2020a2d2,
  title     = {A2D2: Audi Autonomous Driving Dataset},
  author    = {Geyer, Christian and Kassahun, Yarin and Mahmudi, Mulugeta and Ricou, Xavier and Durgesh, Ramkishan and Chung, Shyam and Hauswald, Markus and Pham, Viet and M{\"u}hlethaler, Thomas and Dorn, Sebastian and Fernandez, Ignacio and J{\"a}hne, Bernd and Schmid, Cordelia},
  booktitle = {Proceedings of the IEEE International Conference on Robotics and Automation (ICRA)},
  year      = {2020}
}

@InProceedings{Wang_2023_CVPR,
    author    = {Wang, Chien-Yao and Bochkovskiy, Alexey and Liao, Hong-Yuan Mark},
    title     = {YOLOv7: Trainable Bag-of-Freebies Sets New State-of-the-Art for Real-Time Object Detectors},
    booktitle = {Proceedings of the IEEE/CVF Conference on Computer Vision and Pattern Recognition (CVPR)},
    month     = {June},
    year      = {2023},
    pages     = {7464--7475}
}

@misc{openpcdet2020,
  author       = {OpenPCDet Development Team},
  title        = {OpenPCDet: An Open-Source Toolbox for 3D Object Detection from Point Clouds},
  year         = {2020},
  howpublished = {\url{https://github.com/open-mmlab/OpenPCDet}}
}

@article{Song_2024,
   title={Synthetic Datasets for Autonomous Driving: A Survey},
   volume={9},
   ISSN={2379-8858},
   url={http://dx.doi.org/10.1109/TIV.2023.3331024},
   DOI={10.1109/tiv.2023.3331024},
   number={1},
   journal={IEEE Transactions on Intelligent Vehicles},
   publisher={Institute of Electrical and Electronics Engineers (IEEE)},
   author={Song, Zhihang and He, Zimin and Li, Xingyu and Ma, Qiming and Ming, Ruibo and Mao, Zhiqi and Pei, Huaxin and Peng, Lihui and Hu, Jianming and Yao, Danya and Zhang, Yi},
   year={2024},
   month=jan, pages={1847–1864} }

@inproceedings{talwar_et_al,
author = {Talwar, Deepak and Guruswamy, Sachin and Ravipati, Naveen and Eirinaki, Magdalini},
year = {2020},
month = {08},
pages = {73-80},
title = {Evaluating Validity of Synthetic Data in Perception Tasks for Autonomous Vehicles},
doi = {10.1109/AITEST49225.2020.00018}
}

@article{yolo_v3,
  author       = {Joseph Redmon and
                  Ali Farhadi},
  title        = {YOLOv3: An Incremental Improvement},
  journal      = {CoRR},
  volume       = {abs/1804.02767},
  year         = {2018}
}

@misc{bai2023ffhquvnormalizedfacialuvtexture,
      title={FFHQ-UV: Normalized Facial UV-Texture Dataset for 3D Face Reconstruction}, 
      author={Haoran Bai and Di Kang and Haoxian Zhang and Jinshan Pan and Linchao Bao},
      year={2023},
      eprint={2211.13874},
      archivePrefix={arXiv},
      primaryClass={cs.CV},
      url={https://arxiv.org/abs/2211.13874}, 
}

@misc{goodfellow2014generativeadversarialnetworks,
      title={Generative Adversarial Networks}, 
      author={Ian J. Goodfellow and Jean Pouget-Abadie and Mehdi Mirza and Bing Xu and David Warde-Farley and Sherjil Ozair and Aaron Courville and Yoshua Bengio},
      year={2014},
      eprint={1406.2661},
      archivePrefix={arXiv},
      primaryClass={stat.ML},
      url={https://arxiv.org/abs/1406.2661}, 
}

@inproceedings{karras2019stylegan,
  author    = {Tero Karras and
               Samuli Laine and
               Timo Aila},
  title     = {A Style-Based Generator Architecture for Generative Adversarial Networks},
  booktitle = {Proceedings of the IEEE Conference on Computer Vision and Pattern Recognition (CVPR)},
  year      = {2019}
}

@article{FID_DBLP:journals/corr/HeuselRUNKH17,
  author       = {Martin Heusel and
                  Hubert Ramsauer and
                  Thomas Unterthiner and
                  Bernhard Nessler and
                  G{\"{u}}nter Klambauer and
                  Sepp Hochreiter},
  title        = {GANs Trained by a Two Time-Scale Update Rule Converge to a Nash Equilibrium},
  journal      = {CoRR},
  volume       = {abs/1706.08500},
  year         = {2017}
}

@misc{clip_radford2021learningtransferablevisualmodels,
      title={Learning Transferable Visual Models From Natural Language Supervision}, 
      author={Alec Radford and Jong Wook Kim and Chris Hallacy and Aditya Ramesh and Gabriel Goh and Sandhini Agarwal and Girish Sastry and Amanda Askell and Pamela Mishkin and Jack Clark and Gretchen Krueger and Ilya Sutskever},
      year={2021},
      eprint={2103.00020},
      archivePrefix={arXiv},
      primaryClass={cs.CV},
      url={https://arxiv.org/abs/2103.00020}, 
}

@inproceedings{2022role,
  title={The role of imagenet classes in Fr\'echet inception distance},
  author={Kynk{\"a}{\"a}nniemi, Tuomas and Karras, Tero and Aittala, Miika and Aila, Timo and Lehtinen, Jaakko},
  booktitle={International Conference on Learning Representations (ICLR)},
  year={2023}
}

@inproceedings{precisionrecallkynkaanniemi2019improved,
  title={Improved precision and recall metric for assessing generative models},
  author={Kynk{\"a}{\"a}nniemi, Tuomas and Karras, Tero and Laine, Samuli and Lehtinen, Jaakko and Aila, Timo},
  booktitle={Neural Information Processing Systems (NeurIPS)},
  year={2019}
}

@inproceedings{he2016resnet,
  title={Deep residual learning for image recognition},
  author={He, Kaiming and Zhang, Xiangyu and Ren, Shaoqing and Sun, Jian},
  booktitle={Conference on Computer Vision and Pattern Recognition (CVPR)},
  year={2016}
}

@article{ma2015chicago,
  title={The Chicago face database: A free stimulus set of faces and norming data},
  author={Ma, Debbie S and Correll, Joshua and Wittenbrink, Bernd},
  journal={Behavior research methods},
  volume={47},
  number={4},
  pages={1122--1135},
  year={2015},
  publisher={Springer}
}

@misc{scucchia2025gamingresearchgtav,
      title={From Gaming to Research: GTA V for Synthetic Data Generation for Robotics and Navigations}, 
      author={Matteo Scucchia and Matteo Ferrara and Davide Maltoni},
      year={2025},
      eprint={2502.12303},
      archivePrefix={arXiv},
      primaryClass={cs.CV},
      url={https://arxiv.org/abs/2502.12303}, 
}

@misc{gaidon2016virtualworldsproxymultiobject,
      title={Virtual Worlds as Proxy for Multi-Object Tracking Analysis}, 
      author={Adrien Gaidon and Qiao Wang and Yohann Cabon and Eleonora Vig},
      year={2016},
      eprint={1605.06457},
      archivePrefix={arXiv},
      primaryClass={cs.CV},
      url={https://arxiv.org/abs/1605.06457}, 
}

@misc{cabon2020virtualkitti2,
      title={Virtual KITTI 2}, 
      author={Yohann Cabon and Naila Murray and Martin Humenberger},
      year={2020},
      eprint={2001.10773},
      archivePrefix={arXiv},
      primaryClass={cs.CV},
      url={https://arxiv.org/abs/2001.10773}, 
}

@InProceedings{Ros_2016_CVPR,
author = {Ros, German and Sellart, Laura and Materzynska, Joanna and Vazquez, David and Lopez, Antonio M.},
title = {The SYNTHIA Dataset: A Large Collection of Synthetic Images for Semantic Segmentation of Urban Scenes},
booktitle = {Proceedings of the IEEE Conference on Computer Vision and Pattern Recognition (CVPR)},
month = {June},
year = {2016}
}

@inproceedings{Szegedy2016InceptionV3,
  title     = {Rethinking the Inception Architecture for Computer Vision},
  author    = {Szegedy, Christian and Vanhoucke, Vincent and Ioffe, Sergey and Shlens, Jonathon and Wojna, Zbigniew},
  booktitle = {Proceedings of the IEEE Conference on Computer Vision and Pattern Recognition (CVPR)},
  year      = {2016}
}

@inproceedings{oehrisynth,
  title={{GenFormer} – Generated Images are All You Need to Improve Robustness of Transformers on Small Datasets},
  author={Oehri, Sven and Ebert, Nikolas and Abdullah, Ahmed and Stricker, Didier and Wasenm{\"u}ller, Oliver},
  booktitle={International Conference on Pattern Recognition (ICPR)},
  year={2024}
}

@inproceedings{fang2024data,
  title={Data augmentation for object detection via controllable diffusion models},
  author={Fang, Haoyang and Han, Boran and Zhang, Shuai and Zhou, Su and Hu, Cuixiong and Ye, Wen-Ming},
  booktitle={Winter Conference on Applications of Computer Vision (WACV)},
  year={2024}
}

@inproceedings{abdullahboosting,
    author    = {Abdullah, Ahmed and Ebert, Nikolas and Wasenm\"uller, Oliver},
    title     = {Boosting Few-Shot Detection with Large Language Models and Layout-to-Image Synthesis},
    booktitle = {Asian Conference on Computer Vision (ACCV)},
    year      = {2024},
}
